\documentclass[letterpaper, 10 pt, conference]{ieeeconf}  

\makeatletter
\let\NAT@parse\undefined
\makeatother
\usepackage[colorlinks,urlcolor=blue,linkcolor=blue,citecolor=green]{hyperref}
\usepackage{bookmark}
\usepackage{color,array}
\usepackage{cite}
\usepackage{graphicx}
\usepackage{upgreek}
\usepackage{booktabs}  
\usepackage{threeparttable}  
\usepackage{multirow}
\usepackage{epstopdf}
\usepackage{amsmath}
\usepackage{amssymb}
\usepackage{algorithm}
\usepackage{algorithmicx}
\usepackage{algpseudocode}
\usepackage{setspace}
\usepackage{url}
\usepackage{subfig}
\usepackage{gensymb}
\usepackage{soul}
\usepackage{threeparttable}
\soulregister\cite7 
\soulregister\cite7 
\soulregister\citet7 

\IEEEoverridecommandlockouts                              

\title{\LARGE \bf
GKNet: Graph-based Keypoints Network for Monocular Pose Estimation of Non-cooperative Spacecraft
}

\author{Weizhao Ma$^{1}$, Dong Zhou$^{1, *}$, Yuhui Hu$^{1}$ and Zipeng He$^{2}$
\thanks{This work was supported by the National Natural Science Foundation of China (Grant No. 62403162).}
\thanks{$^{1}$Weizhao Ma, Dong Zhou, and Yuhui Hu are with the Department of Control Science and Engineering, Harbin Institute of Technology, Harbin, China.}
\thanks{$^{2}$Zipeng He is with China Academy of Space Technology, Beijing, China.}
\thanks{$^{*}$Correspoding author: Dong Zhou({\tt\small dongzhou@hit.edu.cn})}%
}

\begin{document}

\maketitle
\thispagestyle{empty}
\pagestyle{empty}

\newcommand{\tabincell}[2]{\begin{tabular}{@{}#1@{}}#2\end{tabular}}

\begin{abstract}
 Monocular pose estimation of non-cooperative spacecraft is significant for on-orbit service (OOS) tasks, such as satellite maintenance, space debris removal, and station assembly. Considering the high demands on pose estimation accuracy, mainstream monocular pose estimation methods typically consist of keypoint detectors and PnP solver. However, current keypoint detectors remain vulnerable to structural symmetry and partial occlusion of non-cooperative spacecraft. To this end, we propose a graph-based keypoints network for the monocular pose estimation of non-cooperative spacecraft, GKNet, which leverages the geometric constraint of keypoints graph. In order to better validate keypoint detectors, we present a moderate-scale dataset for the spacecraft keypoint detection, named SKD, which consists of 3 spacecraft targets, 90,000 simulated images, and corresponding high-precise keypoint annotations. Extensive experiments and an ablation study have demonstrated the high accuracy and effectiveness of our GKNet, compared to the state-of-the-art spacecraft keypoint detectors. The code for GKNet and the SKD dataset is available at \url{https://github.com/Dongzhou-1996/GKNet}.
\end{abstract}


\section{INTRODUCTION\label{section1}}
In recent years, rapid advancements in space technology have enabled the deployment of numerous spacecraft for diverse missions, from Earth observation to deep-space exploration. However, numerous spacecraft have been rendered non-cooperative space debris due to propellant depletion and technical malfunctions. The proliferation of such non-cooperative spacecraft has exacerbated congestion in critical orbital regions, heightening risks to future missions and jeopardizing the operational safety of active satellites. To safeguard the space environment, national space agencies are actively developing OOS technologies capable of performing refueling, maintenance, and deorbit removal of defunct spacecraft. Central to these missions is the acquisition of the accurate real-time pose (position and orientation) of the non-cooperative spacecraft~\cite{zhou2022space}, which is essential for enabling safe proximity operations and ensuring mission success.  

In contemporary aerospace systems, sensors for pose estimation of non-cooperative spacecraft primarily fall into two categories~\cite{zhang2024dfti}: cameras and LiDAR (Light Detection and Ranging). While LiDAR provides accurate depth measurements, it suffers from significantly higher power consumption and lower spatial resolution compared to optical sensors~\cite{li2020pose}. In contrast, monocular cameras offer distinct advantages, including lower energy requirements, simpler maintenance protocols, and higher resolution imaging, making them particularly well-suited for OOS missions where efficiency and reliability are critical. For monocular pose estimation of non-cooperative spacecraft, existing methodologies are broadly divided into traditional approaches (e.g., marker-based or model-based geometric solutions) and deep learning (DL)-based techniques~\cite{park2022speed}, which leverage neural networks to directly estimate the pose from a monocular image.

Traditional methods rely on hand-engineered feature extraction algorithms, such as SIFT and SURF, and priori knowledge of the spacecraft target \cite{zhang2018vision, sharma2016comparative}. These methods establish 2D-3D correspondences by detecting and matching keypoints, and then solve the Perspective-n-Points (PnP) problem to obtain the pose of non-cooperative spacecraft. However, these algorithms struggle to handle challenging situations, such as complex lighting conditions and occluded features.

\begin{figure*}[t]
    \centering
    \includegraphics[width=0.85\textwidth]{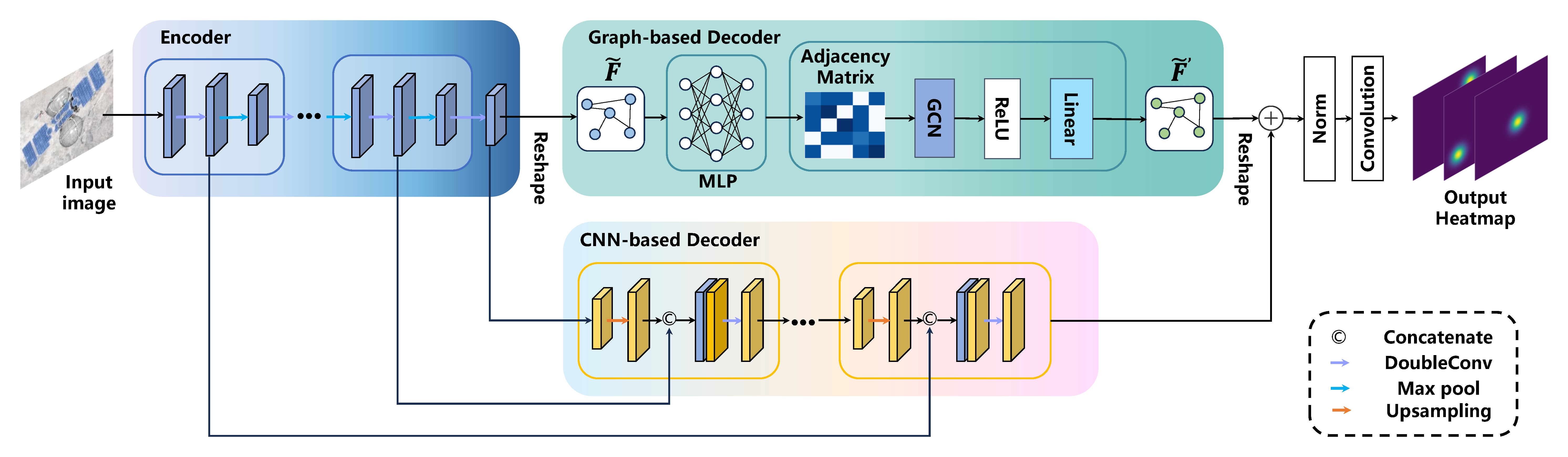} 
    \caption{The overview architecture of GKNet\centering}\label{fig1}
    \centering
 \end{figure*}

The recent advancement of deep learning algorithms, particularly convolutional neural networks (CNNs), has revolutionized pose estimation for non-cooperative spacecraft by enabling automated feature extraction of critical components such as solar panel junctions and thruster clusters~\cite{proencca2020deep, posso2022mobile, garcia2021lspnet}. DL-based algorithms eliminate manual feature engineering, while achieving high pose estimation accuracy through hierarchical representation learning from synthetic datasets. Meanwhile, these approaches demonstrate superior robustness in challenging orbital conditions, such as high illumination variance and thermal-induced image distortions. The approaches of DL-based spacecraft pose estimation can be broadly categorized into two types~\cite{pauly2023survey}: direct end-to-end approaches and hybrid modular approaches. Direct end-to-end approaches use a single model to directly regress pose information from spacecraft images. The latter approaches use neural network to detect the keypoints from spacecraft images, and then utilize PnP algorithms to figure out the target's pose. 

Conventional hybrid approaches in spacecraft pose estimation typically process keypoints as isolated features without modeling their structural context. To this end, our GKNet leverages the geometric constraint of keypoints graph, which enables spatial relationship reasoning, high occlusion resilience and symmetry disambiguation of keypoint detectors. The overall architecture of GKNet is illustrated in Fig.~\ref{fig1}. We employ an encoder to extract high-dimensional feature representations from the input image, followed by two parallel decoder branches: an upsampling-based decoder branch and a graph-convolution-based decoder branch. Finally, the outputs of these two branches are fused to produce the final output. Meanwhile, we present the SKD dataset for the training and testing of different keypoint detectors, which consists of 3 spacecraft targets, 90,000 simulated images, and corresponding high-precise keypoint annotations. Experimental results demonstrate that our GKNet achieves high performance on keypoint detection and pose estimation of non-cooperative spacecraft, compared to state-of-the-art methods~\cite{chen2019satellite, cosmas2020utilization}.

The structure of this paper is as follows: Section 2 introduces an overview of related work on spacecraft pose estimation. Section 3 presents the proposed approach. In Section 4, we detail and analyze the experimental results.

\section{Related Work\label{section2}}
\subsection{Direct End-to-end Approaches}

   Direct end-to-end approaches simplify the pose estimation process by eliminating intermediate steps, such as feature extraction, keypoint detection, and 2D-3D matching. These methods often utilize a powerful neural network to directly learn the mapping relationship between the input image and target pose through millions of times training on large-scale pose estimation datasets. 
   
   In general, end-to-end models are more elegant and streamlined than hybrid modular approaches. Proença et al.~\cite{proencca2020deep} proposed URSONet that uses ResNet as the backbone and two branches to predict the orientation and position respectively. The position branch uses two fully connected layers to regress position coordinates. For orientation estimation, they designed a continuous orientation estimation method based on classification and soft-assignment coding. Based on URSONet, Poss et al.~\cite{posso2022mobile} proposed Mobile-URSONet, which replaces the ResNet backbone with MobileNet-v2. It is a more lightweight pose estimation network with negligible performance degradation. Meanwhile, Albert et al.~\cite{garcia2021lspnet} also presented the LSPNet, which first predicts the position and employs the predicted position to support later orientation estimation.

   However, direct end-to-end approaches are more prone to overfitting and require a large amount of data for training. In addition, the pose estimation accuracy of these methods is often lower by an order of magnitude than hybrid modular approaches.

\subsection{Hybrid Modular Approaches}

    Hybrid modular approaches typically utilize a DL-based model to predict keypoints from monocular images, followed by a PnP solver to estimate the pose of spacecraft. This pipeline allows the hybrid modular approaches to achieve higher accuracy in pose estimation, benefiting from the robustness of deep learning while maintaining the precision provided by the PnP solver. Therefore, the core of the hybrid modular approaches lies in keypoint detectors, which is mainly divided into two categories: 

    \textbf{Regression of keypoint coordinates}: Directly regressing keypoint coordinates of non-cooperative spacecraft from input images is a common approach. Wang et al.~\cite{wang2022spacenet} proposed CA-SpaceNet, a model having three FPNs with two DarkNet-53 networks as the backbones. In addition, CA-SpaceNet introduced counterfactual analysis to address the impact of complex backgrounds. Chen et al.~\cite{chen2019satellite} used Faster-RCNN~\cite{ren2015faster} with HRNet-W18C as the backbone for object detection. Subsequently, they utilized Pose-HRNetW32 for keypoint prediction. Meanwhile, Park et al.~\cite{park2019towards} proposed a different approach, which utilized a YOLOv2-based architecture~\cite{redmon2017yolo9000} with MobileNetv2 as the backbone to directly predict keypoint coordinates. 

    \textbf{Regression of heatmaps}: Another type of keypoint detectors for non-cooperative spacecraft is to regress heatmaps that reflect the probability of keypoint locations. The keypoint coordinates are then obtained by calculating the locations with the highest probability in the heatmaps. The ground truth heatmaps are generated as 2D Gaussian distributions, of which mean values are the coordinate of ground-truth keypoints. Compared to directly regressing keypoint coordinates, this kind of method typically achieves higher accuracy and better generalization. In recent, representative architectures like High-Resolution Net (HRNet) and U-Net have become prevalent for keypoint heatmaps prediction. HRNet maintains high-resolution representations through parallel multi-scale convolutions while preserving spatial precision, whereas U-Net employs symmetric encoder-decoder structure with skip connections to capture multi-scale contextual features. Cosmas et al.~\cite{cosmas2020utilization} combined the ResNet34 with U-Net to regress heatmaps. Huo et al.~\cite{huo2020fast} also proposed a novel lightweight YOLO-liked CNN to predict keypoints with heatmaps.

    However, the existing keypoint detectors typically process keypoints as isolated features without modeling their structural context. To this end, our GKNet leverages the geometric constraint of keypoints graph, which enables spatial relationship reasoning, high occlusion resilience and symmetry disambiguation.

\section{Methods}
    Formally, the goal of this study is to predict the pose of spacecraft from a monocular image. We propose the GKNet (as shown in Fig. \ref{fig1}) to predict the keypoint coordinates, which leverages the geometric constraint of keypoints graph. To better validate keypoint detectors, we present a moderate-scale dataset for the spacecraft keypoint detection, named SKD, which consists of 3 spacecraft targets, 90,000 simulated images, and corresponding high-precise keypoint annotations.
 
    \subsection{GKNet}

    Our GKNet adopts an UNet-like architecture, which was originally developed for image segmentation tasks, consisting of downsampling and upsampling. In the downsampling process, the GKNet utilizes five downsampling layers to extract the deeper feature $F$ from the input image $I$. Then, $F$ is processed through two branches: upsampling branch and graph convolution network (GCN) branch. 
    
    The upsampling branch adjusts $ F $ to the size of the input image $I$, resulting in the feature $F^{\prime}$. Subsequently, we employ a convolution operation to transform $F^{\prime}$ into $F^{\prime\prime} \in \mathbb{R} ^{H \times W \times N}$. This process is expressed as:
    \begin{equation} \label{eq:equ1}
       {F^{\prime\prime}=Conv(Up(F))},
    \end{equation}
    where $H \times W$ is the size of the target heatmap and $N$ is the number of keypoints.
 
    Meanwhile, we incorporate GCN into the GKNet. In general, GCN is not suitable for processing simple graph structures, as they may lead to excessive similarity between features, resulting in an over-smoothing issue~\cite{oono2019graph}. To solve this problem, as shown in Fig.~\ref{fig1}, we introduce a linear layer after each graph convolution, following GraphCape~\cite{hirschorn2024graph}, to form the graph-based decoder.
 
    In the graph-based decoder, $F$ is transformed from a 2D tensor into a 1D vector $\tilde{F} \in \mathbb{R} ^{N \times C_{\text{in}}}$, with $C_{\text{in}}$ representing the feature dimension of each keypoint. Next, we apply the graph-based decoder to process the keypoint features. The output feature $\tilde{F'}\in \mathbb{R} ^{N \times C_{\text{out}}}$ can be formulated as:
    \begin{equation} \label{eq:equ2}
       {\tilde{F'} = \sigma \left(W_{\text{adj}} \tilde{F} \tilde{A}\right)},
    \end{equation}
   where $W_{\text{adj}}\in \mathbb{R} ^{C_{\text{out}} \times C_{\text{in}}}$ is a learnable parameter matrix, $\sigma$ is an activation function (ReLU), and $\tilde{A} \in \mathbb{R} ^{N \times N}$ is the symmetrically normalized form of the adjacency matrix $A \in [0, 1]^{N \times N}$. $A$ is a binary matrix, where the elements are defined as follows:
    \begin{equation} \label{eq:equ3}
       {
          a_{ij} = \begin{cases} 
       1 & \text{if node } v_j \text{ is connected to node } v_i, \\
       1 & \text{elif } i = j, \\
       0 & \text{else}
       \end{cases}
       }
    \end{equation}
    The connectivity between nodes is defined based on a combination of spatial proximity and semantic relationships. Finally, $F''$ and $\tilde{F'}$ are fused, followed by a $ 1 \times 1 $ convolutional layer to obtain the target heatmap $M \in \mathbb{R} ^{N \times H \times W}$. 
 
    Following HRNet~\cite{sun2019deep}, we employ heatmap loss to supervise model training. The loss function is defined as the MSE (Mean-Square Error), which calculates the pixel-level loss and can be expressed as follows:
    \begin{equation} \label{eq:equ4}
       {\mathcal{L}= \frac{1}{W \times H} \sum_{w=1}^{W} \sum_{h=1}^{H} \left( M_{w,h} - \hat{M}_{w,h} \right)^2},
    \end{equation}
    where $\hat{M}\in \mathbb{R} ^{N \times H \times W}$ represents the ground truth heatmap generated from keypoints annotations $\hat{k} =\{(\hat{x}_i, \hat{y}_i) \vert i=1,2, \ldots, N\}$.

    It is worth noting that once the GKNet is trained well, the keypoints of non-cooperative spacecraft $k=\{(x_i, y_i) \vert i = 1,2, \dots, N \}$ can be obtained by calculating the locations with the highest probability in the heatmap, which can be formulated as:
    \begin{equation}
      (x_i, y_i) = \lambda * \arg \max_{w, h} M_{w, h}^{i}
    \end{equation}
    in which, $\lambda$ is the equivalent scale ratio of the GKNet, which is used to convert the heatmap coordinates to the original image coordinates.

   \begin{figure*}[!ht]
      \centering
      \begin{tabular}{p{0.35\textwidth}<{\centering} p{0.4\textheight}<{\centering}}
      \subfloat[Satellite 01]{\includegraphics[width=0.3 \textwidth, height=0.12\textheight]{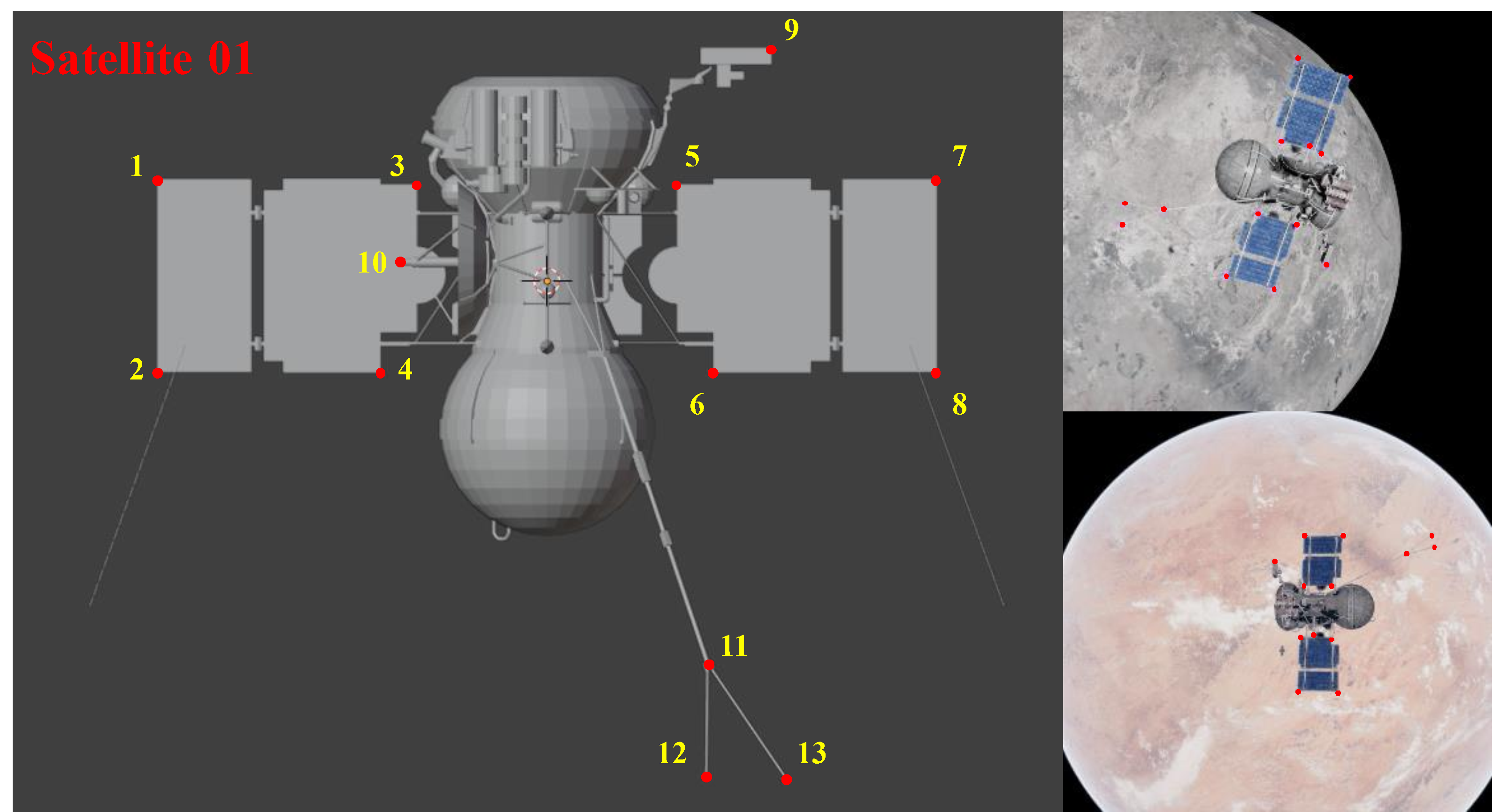}\label{fig2_a}} &
      \multirow{2}{*}[-1em]{\subfloat[Satellite 02]{\includegraphics[width = 0.48\textwidth, height=0.28\textheight]{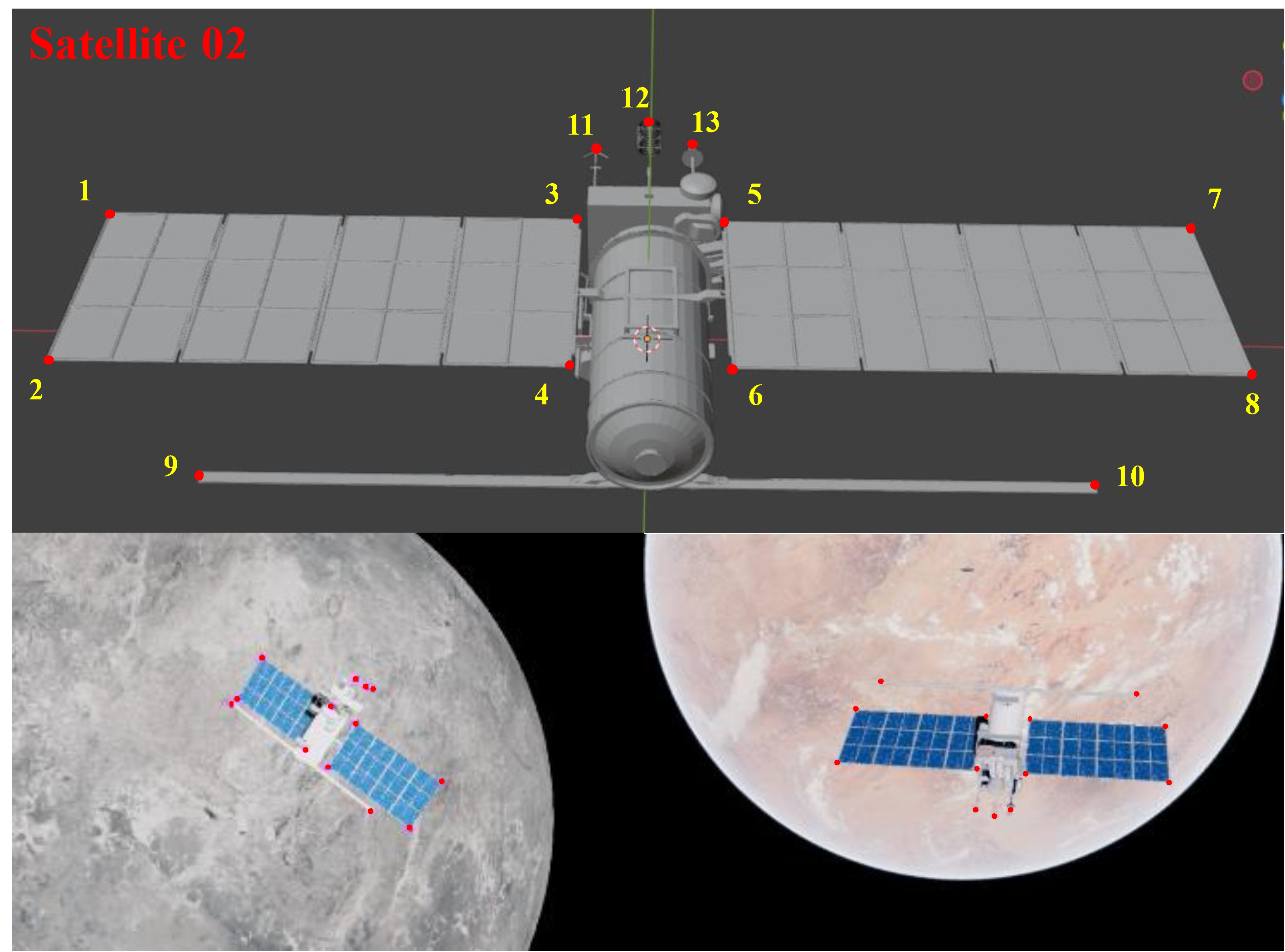}\label{fig2_b}}} \\
      \subfloat[Satellite 03]{\includegraphics[width=0.3 \textwidth, height=0.12\textheight]{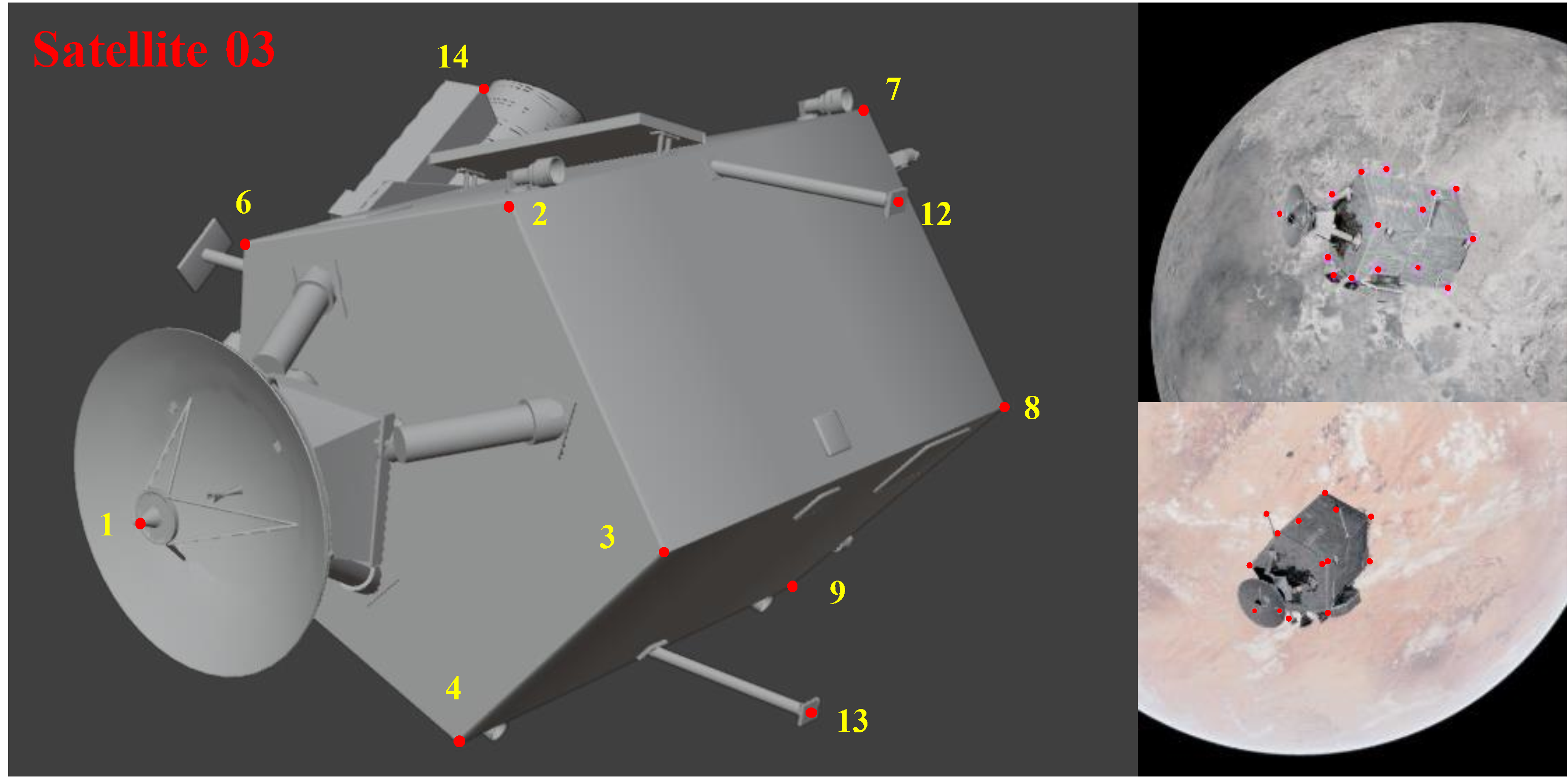}\label{fig2_c}} & \\
      \end{tabular}
      \caption{The keypoints definition of three spacecraft types and the visualization of the SKD dataset. \centering}\label{fig2}
   \end{figure*}

   \subsection{SKD Dataset}
   
   The SKD dataset consists of 3 spacecraft targets (i.e., Satellite 01, Satellite 02, and Satellite 03), as shown in Fig.~\ref{fig2}. According to the discrimination of important components features (e.g. solar panels, thrusters, and antennas), we predefine 10+ keypoints for each spacecraft target and measure the 3D coordinates of keypoints $\hat{K}^b =\{(\hat{x}_i^b, \hat{y}_i^b, \hat{z}_i^b) \vert i=1,2, \ldots, N\}$ in the body frame. Therefore, we can generate high-precise 2D keypoints annotations $\hat{k}^c =\{(\hat{x}_i^c, \hat{y}_i^c) \vert i=1,2, \ldots, N\}$ in the camera frame by camera projection principle:
   \begin{equation}
      \begin{bmatrix}
         \hat{x}_i^c \\
         \hat{y}_i^c
      \end{bmatrix}
      = \begin{bmatrix}
         f_x & 0 & c_x \\
         0 & f_y & c_y \\
         0 & 0 & 1
      \end{bmatrix}
      \begin{bmatrix}
         R_{bc} & T_{bc} \\
      \end{bmatrix}
      \begin{bmatrix}
         \hat{x}_i^b \\
         \hat{y}_i^b \\
         \hat{z}_i^b \\
         1
      \end{bmatrix}
   \end{equation}
   in which, $f_x$ and $f_y$ are the focal lengths of the virtual camera, $c_x$ and $c_y$ are the principal points, $R_{bc}$ and $T_{bc}$ are the rotation matrix and translation vector from the body frame to the camera frame, respectively.

   To generate more realistic data under diverse visual conditions, we import those three spacecraft targets into the Unreal Engine 4 (UE4) platform and render the simulated images with accurate keypoints and pose annotations. In final, we collect 300 simulated video sequences to compose the SKD dataset, of which length is 300 frames, and each frame has a resolution of 1024×1024. The dataset is divided by a ratio of 3:1:1 into three parts: train, val, and test set. Part of the dataset is shown in Fig.~\ref{fig2}.

\begin{figure*}[t]
    \centering
    \begin{minipage}{0.7\textwidth} 
        \centering
        \includegraphics[width=\linewidth,keepaspectratio]{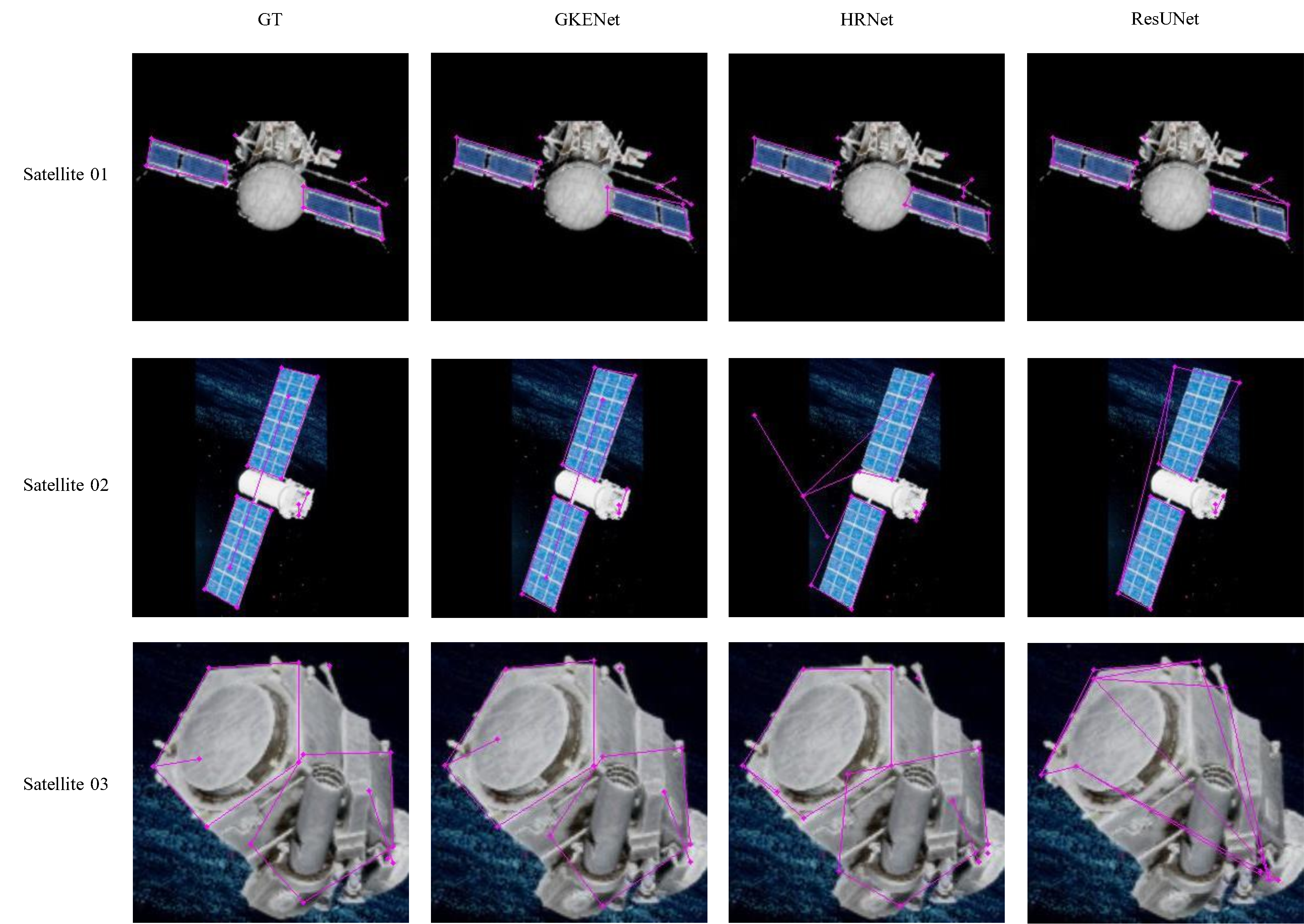} 
        \caption{Keypoint detection results of different keypoint detectors on the SKD dataset}
        \label{fig3}
    \end{minipage}
\end{figure*}

   \section{Experiments}
   
   In this section, we evaluate our GKNet on the SKD dataset compared with other SOTA methods~\cite{chen2019satellite, cosmas2020utilization} which are trained from scratch by ourselves. Meanwhile, an ablation study is implemented to show the effectiveness of our method. All the experiments were conducted on high performance computer server equipped with Intel Xeon Gold 6132 CPU and Nvidia Tesla P100 GPU.

\subsection{Implementation Details}

  We train GKNet on the train set, use val set to adjust parameters and validate performance, and finally evaluate the model on the test set. The training procedures are conducted repetitively for each spacecraft target.

  The trainings of different algorithms all share the following parameters:
   \begin{enumerate}
      \item The original image is cropped based on its keypoint annotations and then uniformly padded and resized to 256×256.
      \item Batch size of 16 samples for 80 epochs.
      \item Adam optimizer with an initial learning rate of 0.01.
      \item Cosine annealing learning rate schedule is used to prevent the model from getting trapped in local minima, where the learning rate gradually decays to 0 over the course of 16 epochs.
      \item Output heatmap size is 32×32.
   \end{enumerate}

\subsection{Keypoint Detection Results}

  We first use Root Mean Square Error (RMSE) as the metric to evaluate the three keypoint detectors on SKD dataset, of which results have been summarized in Table~\ref{table1}:

   \begin{equation} \label{eq:equ5}
      {
   \text{RMSE} = \sqrt{\frac{1}{N} \sum_{i=1}^{N} \left( (x_i - \hat{x}_i)^2 + (y_i - \hat{y}_i)^2 \right)}
   },
   \end{equation}
   where $(x_i, y_i)$ denotes the coordinate of detected keypoint, $(\hat{x}_i, \hat{y}_i)$ represents the corresponding ground-truth, and $N$ is the number of keypoints predefined for each spacecraft target.

\begin{table}[!t]
    \centering
    \caption{The evaluation results of different keypoint detectors on SKD dataset}
    \label{table1}
    \begin{tabular}{cccc}
         \toprule
        Satellite Type  & HRNet & ResUNet & GKNet \\
        \midrule
        Satellite01  & 6.1205 & 5.6224 & \textbf{5.3830} \\
        Satellite02  & 74.6921 & 59.2225 & \textbf{29.1077} \\
        Satellite03  & 32.6379 & 54.1081 & \textbf{24.5821} \\
        \bottomrule
    \end{tabular}
\end{table}

   As shown in Table~\ref{table1}, our method consistently outperforms HRNet~\cite{chen2019satellite} and ResUNet~\cite{cosmas2020utilization}  across all the non-cooperative spacecraft targets. Meanwhile, the GKNet demonstrates a 2.5x improvement in keypoint detection accuracy over HRNet on the Satellite02 target. Furthermore, our method maintains superior performance on Satellite03, which exhibits significantly different geometric characteristics compared to conventional spacecraft targets. To better show the performance gap between the three keypoint detectors, we also depict their keypoint detection results in the Fig.~\ref{fig3}. It can be clearly shown that the detection results of our method on the SKD dataset are nearly the same as the keypoints annotations. Benefiting from the graph-based decoder, the GKNet can precisely predict those intractable keypoints which have been heavily occluded or lacks of discriminative features. 

   \begin{table}[!t]
      \centering
      \caption{The pose estimation performances of different keypoint detectors on SKD dataset}\label{table2}
      \begin{tabular}{cccc}
         \toprule
         Targets & Methods & $E_t$ & $E_q$/rad \\ 
         \midrule
         \multirow{4}{*}{\centering Satellite01} & DMANet & 1.1362 & 0.9971 \\
         & HRNet & 0.7536 & 0.6124 \\ 
         & ResUNet & \textbf{0.7247} & 0.5534 \\
         & GKNet & 0.7629 & \textbf{0.5470} \\ 
         \midrule
         \multirow{4}{*}{\centering Satellite02} & DMANet & 2.6273 & 1.8206 \\
         & HRNet & 3.0422 & 2.1571 \\ 
         & ResUNet & 1.4160 & 1.9951 \\
         & GKNet & \textbf{1.1634} & \textbf{1.0164} \\ 
         \midrule
         \multirow{4}{*}{\centering Satellite03} & DMANet & 1.4052 & 2.7451 \\
         & HRNet & 1.1512 & 2.0285 \\ 
         & ResUNet & 1.2807 & 1.9532 \\
         & GKNet & \textbf{0.9782} & \textbf{1.4849} \\ 
         \bottomrule
      \end{tabular}
   \end{table}

\subsection{Pose Estimation Results}
  To further validate the effectiveness of the proposed method for spacecraft pose estimation, we concatenate the GKNet with a conventional PnP solver and evaluate it on the SKD dataset with $E_t$ and $E_q$ as performance metrics:
   \begin{equation} \label{equ6}
   {E_t = \frac{1}{N} \sum_{i=1}^{N} \frac{\| \mathbf{t}_i - \hat{\mathbf{t}}_i \|_2}{\| \hat{\mathbf{t}}_i \|_2}},
   \end{equation}
   where $\mathbf{t}_i$ and $\hat{\mathbf{t}}_i$ are the predicted and ground-truth position vector, respectively.

   \begin{equation} \label{equ7}
   {E_q = \frac{1}{N} \sum_{i=1}^{N} 2 \cdot \arccos \left( \left| \langle \mathbf{q}_i, \mathbf{{q}}_i \rangle \right| \right)},
   \end{equation}
   in which $\mathbf{q}_i$ and $\hat{\mathbf{q}}_i$ are the predicted quaternion and corresponding quaternion annotation.

  Meanwhile, we combine the HRNet~\cite{chen2019satellite} and ResUNet~\cite{cosmas2020utilization} with the same PnP solver, which have been utilized for better comparison of pose estimation performances. And we also incorporate a novel directly end-to-end spacecraft pose estimation method, DMANet~\cite{zhao2024dmanet}, as the baseline. It is worth noting that this method is separately trained from scratch for each spacecraft targets in the SKD dataset. 

  All the evaluation results of the four spacecraft pose estimation algorithms on SKD dataset have been listed in Table~\ref{table2}. It clearly shows the advancement of the GKNet-based pose estimation method, which almost achieves the best $E_t$ and $E_q$ metrics among all the non-cooperative spacecraft targets. In addition, the evaluation results also demonstrate the superiority of hybrid modular pose estimation algorithms, compared to the directly end-to-end approaches.

\subsection{Ablation study}
   To evaluate the effectiveness of our proposed method, we perform an ablation study on the GKNet, in which the Graph-based decoder is removed from the original GKNet, denoted as "w/o GCN". The ablation results are shown in Table~\ref{table3}. The results indicate that the performance of GKNet without graph-based decoder obviously degrades on all three spacecraft targets. The performance degradation became more severe for Satellite 02, because of the high symmetric structure and frequent occlusions. It also demonstrates that our GKNet can effectively capture the spatial relationships between keypoints, leading to improved performance in challenging scenarios. 

 \begin{table}[hb]
   \centering
   \caption{Ablation Study Results}\label{table3}
   \begin{tabular}{cccc}
      \toprule
      Targets & Methods &  RMSE  \\ 
      \midrule
      \multirow{2}{*}{\centering Satellite01} & GKNet & 5.3830  \\ 
      & w/o GCN & 5.9207  \\ 
      \midrule
      \multirow{2}{*}{\centering Satellite02} & GKNet & 29.1077  \\ 
      & w/o GCN & 37.3007  \\ 
      \midrule
      \multirow{2}{*}{\centering Satellite03} & GKNet & 24.5821  \\ 
      & w/o GCN & 27.0278  \\ 
      \bottomrule
   \end{tabular}
\end{table}

\section{Conclusion}
In this paper, we present the GKNet for the monocular pose estimation of non-cooperative spacecraft, which leverages the geometric constraint of keypoints graph. In order to better validate keypoints detectors, we also provide the SKD dataset for the spacecraft keypoints detection, which consists of 3 spacecraft targets, 90,000 simulated images, and corresponding high-precise keypoint annotations. Extensive experiments and the ablation study have demonstrated the high accuracy and effectiveness of our proposed method, compared to the state-of-the-art spacecraft keypoints detectors.






\bibliographystyle{IEEEtran}
\bibliography{GKNet}


\end{document}